\pdfoutput=1
\documentclass[11pt]{article}

\usepackage[final]{acl}

\usepackage{times}
\usepackage{latexsym}

\usepackage[T1]{fontenc}
\usepackage[utf8]{inputenc}

\usepackage{microtype}

\usepackage{inconsolata}

\usepackage{graphicx}

\usepackage{longtable}
\usepackage{xcolor}
\usepackage{colortbl}
\usepackage{multicol}
\usepackage{booktabs}
\usepackage{makecell}
\usepackage{amsmath, amssymb}
\usepackage{multirow}
\usepackage{mathtools}
\usepackage[inline]{enumitem}
\usepackage{subcaption}
\usepackage{float}
\usepackage{graphicx}
\usepackage{upgreek}
\usepackage{seqsplit}
\usepackage{color, soul}
\usepackage{arydshln}
\usepackage{scalerel}

\usepackage{amsmath,amsfonts,bm}

\def\eqref#1{equation~\ref{#1}}
\def\1{\bm{1}}

\def\vx{{\bm{x}}}
\def\vy{{\bm{y}}}

\DeclareMathAlphabet{\mathsfit}{\encodingdefault}{\sfdefault}{m}{sl}
\SetMathAlphabet{\mathsfit}{bold}{\encodingdefault}{\sfdefault}{bx}{n}

\usepackage{caption}
\usepackage{subcaption}
\title{Revisiting In-Context Learning with Long Context Language Models}

\author{
    \textbf{Jinheon Baek}$^1$\thanks{This work was conducted during an internship at Google.} \; 
    \textbf{Sun Jae Lee}$^2$ \; 
    \textbf{Prakhar Gupta}$^2$ \\
    \textbf{Geunseob (GS) Oh}$^2$ \;
    \textbf{Siddharth Dalmia}$^2$ \;
    \textbf{Prateek Kolhar}$^2$ \\
    KAIST$^1$ \;\; Google DeepMind$^2$ \\
    {
        \normalsize
        \texttt{jinheon.baek@kaist.ac.kr} \; \texttt{\{sunjaelee, prakharguptaz, ohgs, pkolhar\}@google.com}
    }
}

\begin{document}
\maketitle

\begin{abstract}

In-Context Learning (ICL) is a technique by which language models make predictions based on examples provided in their input context. Previously, their context window size imposed a limit on the number of examples that can be shown, making example selection techniques crucial for identifying the maximally effective set of examples. However, the recent advent of Long Context Language Models (LCLMs) has significantly increased the number of examples that can be included in context, raising an important question of whether ICL performance in a many-shot regime is still sensitive to the method of sample selection. To answer this, we revisit these approaches in the context of LCLMs through extensive experiments on 18 datasets spanning 4 tasks. Surprisingly, we observe that sophisticated example selection techniques do not yield significant improvements over a simple random sample selection method. Instead, we discover that the advent of LCLMs has fundamentally shifted the challenge of ICL from that of selecting the most effective examples to that of collecting sufficient examples to fill the context window. Specifically, in certain datasets, including all available examples does not fully utilize the context window; however, by augmenting the examples in context with a simple data augmentation approach, we substantially improve ICL performance by 5\%. 

\end{abstract}
\section{Introduction}
In-Context Learning (ICL) has emerged as a powerful paradigm in natural language processing that enables Language Models (LMs) to learn, adapt, and generalize from examples provided within their input context, eliminating the need for extensive training and parameter updates~\cite{GPT3, ICL/1, ICL/2}. However, due to the limited context lengths of earlier LMs (which accommodate only a few thousand tokens), much of previous ICL work has focused on optimizing sample selection strategies~\cite{ICL/KNN/1, ICL/KNN/2, ICL/diversity/1, ICL/diversity/2, ICL/easy/1, ICL/easy/2}. With the advent of Long Context Language Models (LCLMs), which are capable of processing over a million tokens in a single context window, these constraints are significantly relaxed as it enables including a large number of examples to be used in ICL, known as many-shot ICL~\cite{ManyShotICL/1, ManyShotICL/2}. 

\begin{figure*}
    \centering
    \includegraphics[width=0.975\linewidth]{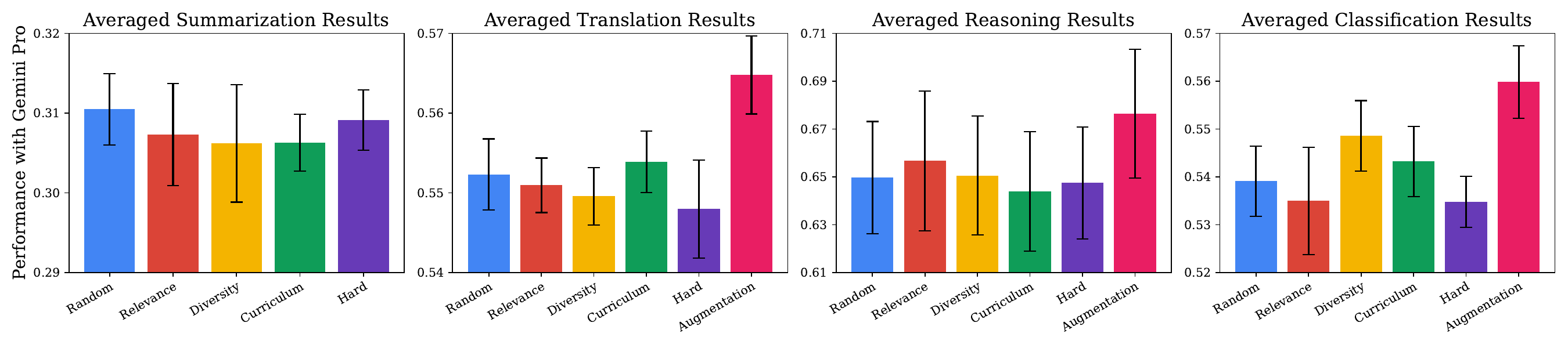}
    \vspace{-0.15in}
    \caption{Results of various sample selection approaches in 64-shot ICL with LCLMs. Approaches include Retrieval that selects examples similar to the target query, Diversity that aims for maximizing example variety, Curriculum that arranges examples in order from easiest to hardest, and Hard that uses only challenging examples, alongside Random that selects examples without any constraints. Results indicate that sample selection methods provide no significant improvement over the naive (random) approach and sometimes perform worse. Meanwhile, Augmentation refers to the approach that generates additional demonstrations and uses them along with original samples for ICL, particularly for low-resource tasks (such as translation, reasoning, and classification) that do not contain enough samples to utilize the full capacity of LCLMs, showing substantial performance gains.}
    \label{fig:overall}
    \vspace{-0.125in}
\end{figure*}

This expansion of context length raises an important question: do previous sample selection strategies, designed for shorter context windows in earlier LMs, generalize to the many-shot ICL regime? To answer this, we systematically revisit existing sample selection strategies by conducting extensive experiments across 18 datasets spanning diverse tasks (namely, classification, translation, summarization, and reasoning) with multiple LCLMs. Our experiments include multiple types of sample selection methods: relevance, diversity, and difficulty-based sample selection, as outlined in~\citet{ICL/survey}. From these experiments, we uncover novel and surprising findings: contrary to prevailing expectations that carefully selected ICL demonstrations would yield performance improvements, they are similarly effective with a simple random selection approach, offering no statistically meaningful improvements in almost all cases (Figure~\ref{fig:overall}). An additional reason to prefer the naive sample selection approach is that it enables greater efficiency through key-value caching of in-context examples (as the same examples can be reused across multiple queries), unlike sophisticated sample selection methods where the examples vary for each sample.

While the expanded context length in LCLMs allows us to focus less on selecting optimal subsets of examples, it introduces a new challenge: effectively utilizing this expanded capacity when the number of examples is limited. Specifically, in scenarios where available data is sparse (such as low-resource translation or reasoning tasks where annotated data samples are difficult or costly to obtain), the examples available only utilize a small fraction of the full context window. This mismatch between context capacity and example availability introduces a new direction in ICL research, shifting the focus from optimizing sample selection to maximally utilizing the long context window. To address this, we propose a simple yet effective data augmentation approach to increase the number of in-context examples, which consists of two steps: (1) generating synthetic examples and (2) filtering out low-quality examples through LCLM prompting contextualized with real examples. Then, by adding these augmented data samples to the context, we significantly improve ICL performance.

Moreover, we explore other key factors unique to LCLM-enabled ICL. Specifically, we investigate the capacity of LCLMs to comprehend extremely long context (where a large number of examples up to the context length are present), as well as how they handle scenarios in which some of these examples introduce noise. Through comprehensive analyses, we find that while performance generally improves as the number of in-context examples increases, it eventually plateaus and begins to decline as the context length approaches the limit. This diminishing return highlights the need to carefully balance context length and example quantity. Also, we observe that LCLMs exhibit robustness to noisy examples in relatively simple tasks, but become vulnerable to noise in more complex scenarios to which they might be less exposed during training, such as extremely low-resource translation tasks.

Overall, we believe our work sheds new light on an important paradigm shift in ICL with LCLMs: the shift from optimizing sample selection to better utilizing extensive context capacity. In particular, our findings suggest that simpler, more efficient random sampling approaches can be as effective as previous sample selection approaches in many-shot settings in most cases, and that data augmentation can significantly improve ICL performance in low-resource tasks. Furthermore, our study paves the way for future research on understanding how to better utilize large context windows and manage the intricacies that arise in extended-context ICL.

\section{Examining Sample Selection Methods for In-Context Learning with LCLMs}

\subsection{Background}
We begin with formally introducing LCLMs, followed by describing the setup of ICL with LCLMs.

\begin{figure*}[t]
    \centering
    \includegraphics[width=0.975\linewidth]{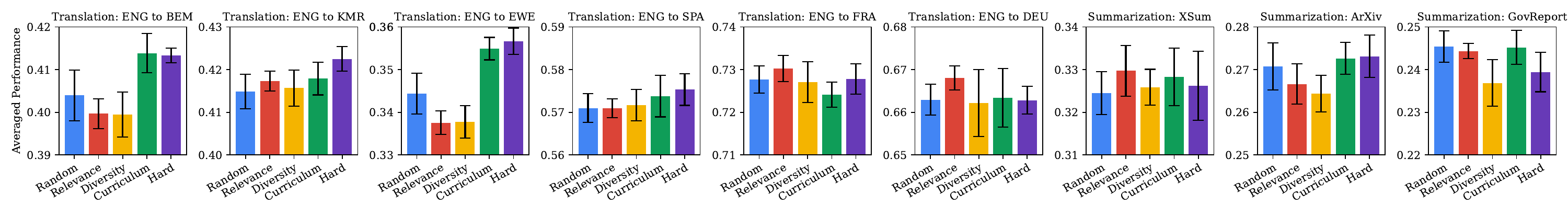}

    \vspace{-0.05in}
    
    \includegraphics[width=0.975\linewidth]{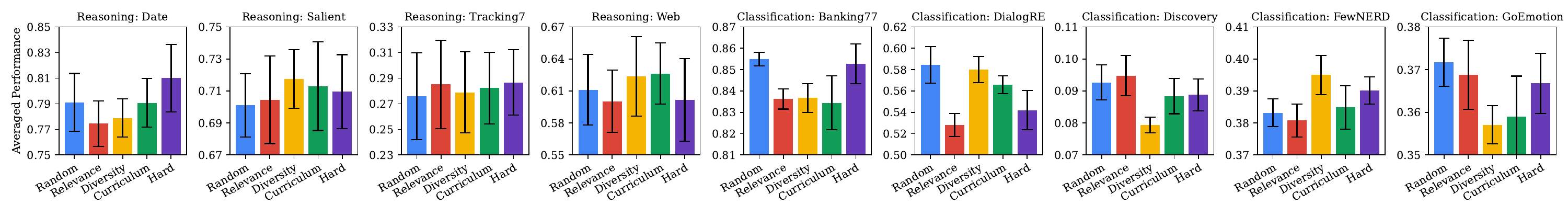}

    \vspace{-0.15in}
    \caption{Results of various sample selection approaches on ICL of 64 examples with LCLMs, where we average the performance over all models: Gemini Pro, Gemini Flash, and Llama 3.1, across four different tasks with 18 datasets. Each bar represents the averaged performance, with the upper and lower limits indicating standard deviation. See Figure~\ref{fig:selection} for results on each model.}
    \label{fig:selection_average}
    \vspace{-0.1in}
\end{figure*}

\paragraph{Long-Context Language Models}
A language model ($\texttt{LM}$), which takes an input sequence of tokens $\vx = [x_1, x_2, \ldots, x_n]$ and generates an output sequence of tokens $\vy = [y_1, y_2, \ldots, y_m]$, can be denoted as follows: $\vy = \texttt{LM}_{\theta}(\vx)$, where $\theta$ is the set of model parameters. A long-context LM ($\texttt{LCLM}$) is an advanced LM~\cite{gemini} that is designed to accommodate sequences with a large number of tokens (e.g., $n$ can exceed 1 million), typically far surpassing the context sizes of earlier LMs.

\paragraph{In-Context Learning with LCLMs}
Given a set of $k$ input-output pairs $\{(\vx_i, \vy_i)\}_{i=1}^k$ as well as an input query $\vx'$, the goal of ICL is to produce an output $\vy = \texttt{LCLM}(\vx' | \{(\vx_i, \vy_i)\}_{i=1}^k)$, where the model ($\texttt{LCLM}$) uses the contextual examples $\{(\vx_i, \vy_i)\}_{i=1}^k$ to make predictions for $\vx'$. In prior research before the advent of LCLMs, the value of $k$ was often limited by the relatively short context lengths of earlier models, which constrained the number of examples that could be utilized for ICL. Subsequently, significant work has focused on developing sample selection techniques to optimize performance within these restricted contexts~\cite{ICL/KNN/1, ICL/KNN/2, ICL/diversity/1, ICL/diversity/2, ICL/easy/1, ICL/easy/2}. In the meantime, the expanded context capacity of LCLMs enables a larger $k$, facilitating many-shot learning with a far greater number of examples.

\subsection{Experimental Setup}
We now discuss the detailed experimental design.

\begin{figure*}
    \centering
    \includegraphics[width=0.99\linewidth]{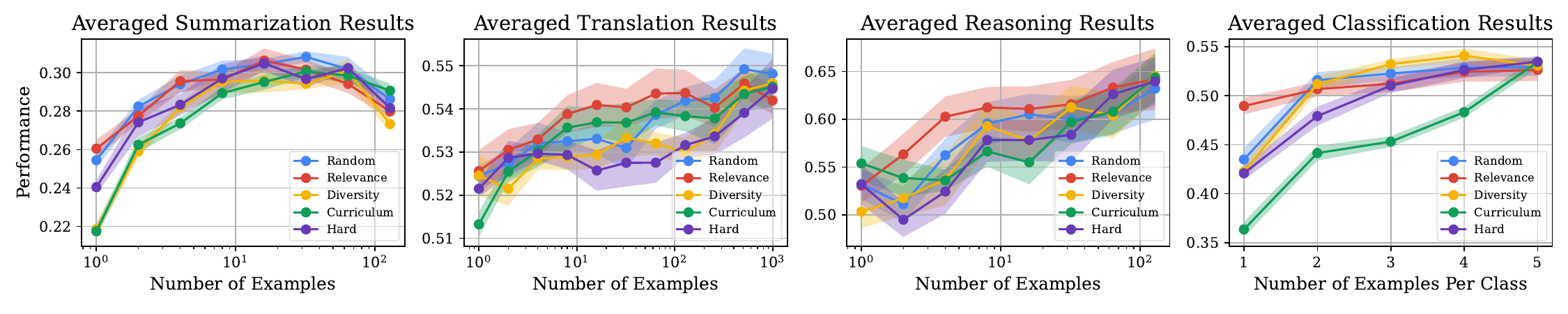}
    \vspace{-0.175in}
    \caption{Results with varying the number of examples for ICL with Gemini Pro, where we average the results for each task.}
    \label{fig:selection_varying_shots}
    \vspace{-0.125in}
\end{figure*}

\paragraph{Tasks and Datasets}
We experiment with 18 different datasets across four tasks to evaluate the effectiveness and robustness of various approaches. 

\vspace{-0.075in}

\begin{itemize}[itemsep=0.75mm, parsep=1pt, leftmargin=*]
    \item \textbf{Translation:} This task evaluates the ability of models to translate text from one language to another. We include translations from English to low-resource languages (namely, Bemba, Northern Kurdish, and Ewe) and high-resource languages (Spanish, French, and German) from the FLORES-200 benchmark~\cite{flores200}, with chrF scores~\cite{chrF} as the metric.
    \item \textbf{Summarization:} This task assesses the capability of models to generate concise and coherent summaries from articles. We include one widely-used XSum dataset~\cite{XSum} and two long-context summarization datasets: ArXiv and GovReport~\cite{ArXiv, GovReport}. ROUGE-L score is used for evaluation.
    \item \textbf{Reasoning:} This task evaluates the ability of models on complex reasoning. We use four challenging datasets from Big Bench Hard~\cite{BBH} following the experimental setting of Long-Context Frontiers (LOFT) benchmark~\cite{loft}, where each data sample follows a multiple-choice question answering format.
    \item \textbf{Classification:} This task includes challenging benchmark datasets for ICL from~\citet{LongICLBench}, particularly designed for classification problems with diverse classes and long inputs.
\end{itemize}

\paragraph{ICL Sample Selection Strategies}
To ensure comprehensive coverage of previously explored sample selection strategies, we follow the category of three core dimensions from~\citet{ICL/survey} (that extensively summarizes around ICL 200 papers). This includes selecting samples based on their diversity, difficulty, and relevance to the query, with the baseline of random sample selection.

\vspace{-0.05in}

\begin{itemize}[itemsep=0.75mm, parsep=1pt, leftmargin=*]
    \item \textbf{Naive:} This method randomly selects examples from a dataset and uses this initial set of selected examples as ICL demonstrations for all queries.
    \item \textbf{Relevance:} This method selects examples that are most similar to the input query to maximize the alignment of ICL demonstrations with the query. To compute semantic similarity between the query and each example, we use the state-of-the-art embedding model~\cite{Gecko}. 
    \item \textbf{Diversity:} This method selects examples that are maximally distinct from each other to capture a broad coverage of features within the task space. We embed each example in a shared embedding space with~\citet{Gecko} and utilize $k$-means clustering (where $k$ corresponds to the number of desired ICL examples) to group the examples into subcategories. We then select the example closest to each cluster center as the representative to capture a diverse subset of the task features.
    \item \textbf{Difficulty:} This method selects examples based on their difficulty. We examine two approaches: the first method (called \textbf{Curriculum}) follows a curriculum learning paradigm where examples are ordered from easiest to hardest; the second one (called \textbf{Hard}) includes only difficult examples, as simpler examples may already be well-understood by models. To assess example difficulty, we use model-based evaluation~\cite{G-Eval} with the state-of-the-art LCLM~\cite{gemini}, which prompts a model 30 times and averages difficulty scores weighted by probabilities. 
\end{itemize}

\begin{table}[t]
\caption{Counting the statistical significance of sophisticated selection approaches over random selection on each experiment instance, by conducting the t-test with 95\% confidence threshold. Tran., Summ., Reas, Clas, denote translation, summarization, reasoning, and classification tasks, respectively.}
\vspace{-0.1in}
\label{tab:significance}
\small
\centering
\resizebox{0.475\textwidth}{!}{
\renewcommand{\arraystretch}{1.0}
\begin{tabular}{llccccc}
\toprule

\textbf{LCLMs} & \textbf{Methods} & \textbf{Tran.} & \textbf{Summ.} & \textbf{Reas.} & \textbf{Clas.} & \textbf{Total} \\

\midrule
\midrule

\multirow{4}{*}{\textbf{Gemini Pro}} 

& Relevance & 0 / 6 & 0 / 3 & 0 / 4 & 0 / 5 & 0 / 18  \\

& Diversity & 0 / 6 & 0 / 3 & 1 / 4 & 2 / 5 & 3 / 18 \\

& Curriculum & 1 / 6 & 0 / 3 & 0 / 4 & 1 / 5 & 2 / 18 \\

& Hard & 0 / 6 & 0 / 3 & 1 / 4 & 0 / 5 & 1 / 18 \\

\midrule

\multirow{4}{*}{\textbf{Gemini Flash}} 

& Relevance & 0 / 6 & 0 / 3 & 0 / 4 & 2 / 5 & 2 / 18  \\

& Diversity & 0 / 6 & 0 / 3 & 0 / 4 & 2 / 5 & 2 / 18 \\

& Curriculum & 0 / 6 & 0 / 3 & 0 / 4 & 0 / 5 & 0 / 18 \\

& Hard & 0 / 6 & 0 / 3 & 0 / 4 & 0 / 5 & 0 / 18 \\

\midrule

\multirow{4}{*}{\textbf{Llama 3.1}} 

& Relevance & 1 / 6 & 0 / 3 & 1 / 4 & 1 / 5 & 3 / 18  \\

& Diversity & 0 / 6 & 0 / 3 & 0 / 4 & 2 / 5 & 2 / 18 \\

& Curriculum & 0 / 6 & 0 / 3 & 0 / 4 & 1 / 5 & 1 / 18 \\

& Hard & 0 / 6 & 0 / 3 & 0 / 4 & 2 / 5 & 2 / 18 \\

\midrule

\multirow{4}{*}{\textbf{Total}} 

& Relevance & 1 / 18 & 0 / 9 & 1 / 12 & 3 / 15 & 5 / 54  \\

& Diversity & 0 / 18 & 0 / 9 & 1 / 12 & 6 / 15 & 7 / 54 \\

& Curriculum & 1 / 18 & 0 / 9 & 0 / 12 & 2 / 15 & 3 / 54 \\

& Hard & 0 / 18 & 0 / 9 & 1 / 12 & 2 / 15 & 3 / 54 \\

\bottomrule

\end{tabular}
}
\vspace{-0.1in}
\end{table}

\paragraph{LCLM Configurations for ICL}
We consider LCLMs that support extensive token capacities to evaluate performance in long-context, many-shot ICL scenarios, such as those with context window lengths on the order of millions: Gemini 1.5 Flash (1M tokens) and Gemini 1.5 Pro (2M tokens)~\cite{gemini}. Also, we consider the Llama 3.1 70B model~\cite{llama3}, which, while supporting the comparatively smaller context size of 128K tokens, is still considered an LCLM. To provide a comprehensive view of performance under different shots, we vary the number of ICL examples, starting from one and sequentially doubling to 2, 4, 8, 16, 32, and so forth, until reaching either the context size limit or the maximum number of dataset samples, whichever is exhausted first. Furthermore, to ensure the reliability of our results, we conduct multiple runs for each setup: 3 runs for translation and summarization tasks and 10 runs for reasoning and classification tasks. The prompts used to elicit responses from ICL are provided in Appendix~\ref{appendix:prompt}.

\subsection{Experimental Results}

\paragraph{Results on Sample Selection Strategies}
We report the detailed results of various sample selection approaches in many-shot ICL scenarios in Figure~\ref{fig:selection_average}. To rigorously evaluate each sample selection approach and their statistically significant gains, we conduct a t-test with a 95\% confidence threshold and report the results in Table~\ref{tab:significance}. From these results, we observe that previously effective sample selection methods, designed for shorter context LMs, yield little to no performance gains over the random selection approach when applied to LCLMs. Aggregated results across three different LCLMs indicate statistical significance in fewer than 15\% of instances, indicating that they are not reliable.

\begin{table}[t]
\caption{Performance comparison of recent sample selection strategies (Auto-ICL, IDS, and ICCL) in many-shot ICL.}
\vspace{-0.1in}
\label{tab:recent_selection}
\small
\centering
\resizebox{0.475\textwidth}{!}{
\renewcommand{\arraystretch}{1}
\begin{tabular}{lcccc}
\toprule

\textbf{Methods} & \textbf{Translation} & \textbf{Summarization} & \textbf{Reasoning} & \textbf{Classification} \\

\midrule
\midrule

Random & 0.551 $\pm$ 0.005 & 0.311 $\pm$ 0.005 & 0.650 $\pm$ 0.020 & 0.539 $\pm$ 0.006 \\

\noalign{\vskip 0.25ex}\cdashline{1-5}\noalign{\vskip 0.75ex}

Auto-ICL & 0.544 $\pm$ 0.003 & 0.305 $\pm$ 0.003 & 0.629 $\pm$ 0.029 & 0.539 $\pm$ 0.005 \\

IDS & 0.547 $\pm$ 0.003 & 0.313 $\pm$ 0.004 & 0.649 $\pm$ 0.018 & 0.537 $\pm$ 0.007 \\

ICCL & 0.553 $\pm$ 0.006 & 0.307 $\pm$ 0.006 & 0.653 $\pm$ 0.016 & 0.543 $\pm$ 0.006 \\

\bottomrule

\end{tabular}
}
\vspace{-0.1in}
\end{table}

\paragraph{Additional Results with Advanced Sample Selection Strategies}
To further assess the robustness of our findings, we additionally evaluate several recent and more advanced sample selection strategies: Auto-ICL~\cite{Auto-ICL}, IDS~\cite{IDS}, and ICCL~\cite{ICL/easy/2}, which have been proposed to improve ICL by selecting high-quality and relevant examples based on context or model feedback. As shown in Table~\ref{tab:recent_selection}, however, we find that none of these newer methods consistently outperform the simple random selection baseline across tasks, with performance fluctuations within the range of statistical variation. This reinforces our main claim that LCLMs are insensitive to the specific sample selection strategy in many-shot ICL.

\begin{figure*}
    \centering
    \includegraphics[width=0.99\linewidth]{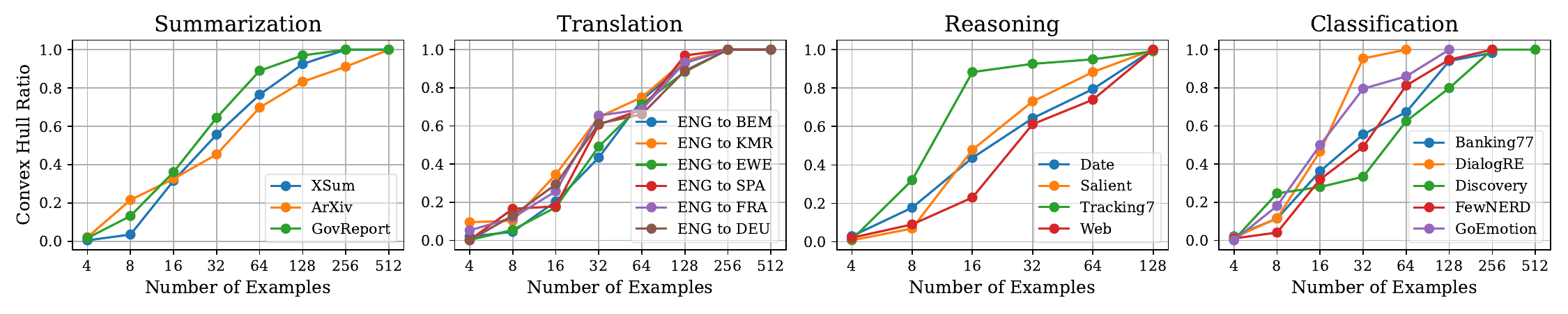}
    \vspace{-0.175in}
    \caption{Ratios of convex hull volume of in-context examples to the full dataset with varying numbers of ICL examples.}
    \label{fig:population_convex}
    \vspace{-0.1in}
\end{figure*}

\paragraph{Analysis on Number of ICL Examples}
To see the performance of ICL with respect to the number of examples, we visualize results in Figure~\ref{fig:selection_varying_shots}. Overall, for any sampling method, we observe that performance increases as the number of examples increases. Also, when the number of examples is relatively small, the relevance-based sample selection approach performs particularly well, as focusing on highly relevant examples maximizes learning effectiveness with the limited number on examples. However, as the number of examples increases, the performance gap between various sample selection methods diminishes, indicating that performance is less dependent on selection strategies in many-shot scenarios. Lastly, in the summarization task (where samples tend to be longer than those in other tasks), we observe an initial increase in performance, followed by a decline once the context becomes heavily populated with a large number of examples. We argue this decline likely reflects the challenges LCLMs face in processing extremely long contexts, discussed in Section~\ref{sec:long_context}.

\paragraph{Analysis on Converge of ICL Examples}
To further investigate why the performance gap between different approaches diminishes as the number of examples increases, we analyze the representational coverage of examples in-context relative to the full examples. Specifically, we measure the convex hull volume spanned by the embeddings of ICL examples (where we vary their numbers) and compare it to that of the entire dataset, which can serve as a proxy for how well the samples in-context capture the distribution of the full data. Our results, visualized in Figure~\ref{fig:population_convex}, show that, when the number of ICL examples is moderate (e.g., 64), they already span over 80\% of the convex hull volume of the full dataset in almost all tasks and datasets. This suggests that, beyond a certain threshold, adding more examples does not significantly improve coverage, as the selected examples, regardless of selection methods, can approximate the full data distribution.

\begin{table}[t]
\caption{Results with varying the order of ICL samples, where Ascending and Descending represent cases where examples closer to the query appear earlier and later in the LCLM context, respectively. In contrast, random denotes the case where examples are arranged randomly without a specific order.}
\vspace{-0.1in}
\label{tab:order}
\small
\centering
\resizebox{0.475\textwidth}{!}{
\renewcommand{\arraystretch}{1}
\begin{tabular}{lcccc}
\toprule

\textbf{Methods} & \textbf{Summarization} & \textbf{Translation} & \textbf{Reasoning} & \textbf{Classification} \\

\midrule
\midrule

Random & 0.310 $\pm$ 0.004 & 0.553 $\pm$ 0.004 & 0.650 $\pm$ 0.023 & 0.539 $\pm$ 0.007 \\

Ascending & 0.307 $\pm$ 0.006 & 0.557 $\pm$ 0.004 & 0.641 $\pm$ 0.027 & 0.534 $\pm$ 0.010 \\

Descending & 0.309 $\pm$ 0.003 & 0.552 $\pm$ 0.007 & 0.648 $\pm$ 0.021 & 0.539 $\pm$ 0.005 \\

\bottomrule

\end{tabular}
}
\vspace{-0.1in}
\end{table}

\paragraph{Analysis on Example Order}
Previous work has shown that earlier LMs are sensitive to the order of examples when doing few-shot ICL. For example, LMs tend to follow the answer in the last example~\cite{order/1, order/2}. To investigate whether similar issues arise in many-shot ICL with LCLMs, we experiment by comparing performance when ordering ICL examples randomly, by increasing similarity, and by decreasing similarity. The results in Table~\ref{tab:order} suggest that the order of examples does not affect performance of LCLMs.

\paragraph{Analysis on Computational Complexity}
In addition to performance, computational complexity is a critical factor to consider when assessing the practicality of many-shot ICL with LCLMs, as they often handle million-token contexts. We note that for approaches that adjust ICL examples based on the given query (such as relevance-based selection), the complexity scales quadratically, $\mathcal{O}(n^2)$, where $n$ represents the number of tokens used for ICL demonstrations. In contrast, the simpler naive selection approach, which uses the same set of randomly selected examples for all queries, offers a significantly more efficient complexity of $\mathcal{O}(kn)$, where $k$ is the number of tokens only within the target query ($n \gg k$). This is because the selected examples do not change based on the query; thus, the same set of examples can be key-value cached. As a result, random selection is a practical choice due to its equivalent performance with other selection methods and the added advantage of efficiency.

\begin{table*}[t]
\caption{Results of LCLM-enabled ICL on four different tasks, where Random indicates the naive sample selection approach without selection criteria, Best Selection indicates the model that achieves the best performance among sophisticated sample selection methods for each experiment unit, and Augmentation indicates the proposed approach that generates demonstrations and uses them alongside original samples with random selection. We emphasize statistically significant results over Random in bold. We exclude Llama from the augmentation scenario as its context capacity is approximately ten times smaller than that of Gemini, allowing it to fully utilize its available context with the original examples alone, making augmentation unnecessary.}
\vspace{-0.1in}
\label{tab:augmentation}
\small
\centering
\resizebox{0.975\textwidth}{!}{
\renewcommand{\arraystretch}{0.975}
\begin{tabular}{llccccccccccccc}
\toprule

& & \multicolumn{6}{c}{Translation} & \multicolumn{2}{c}{Reasoning} \\

\cmidrule(l{2pt}r{2pt}){3-8} \cmidrule(l{2pt}r{2pt}){9-10} 

\textbf{LCLMs} & \textbf{Methods} & ENG to BEM & ENG to KMR & ENG to EWE & ENG to SPA & ENG to FRA & ENG to DEU & Date & Salient \\

\midrule
\midrule

\multirow{3}{*}{\textbf{Gemini Pro}} 

& Random & 0.470 $\pm$ 0.003 & 0.439 $\pm$ 0.001 & 0.419 $\pm$ 0.004 & 0.580 $\pm$ 0.006 & 0.734 $\pm$ 0.002 & 0.676 $\pm$ 0.010 & 0.854 $\pm$ 0.009 & 0.776 $\pm$ 0.035 \\ 

& Best Selection & 0.470 $\pm$ 0.004 & 0.443 $\pm$ 0.004 & 0.418 $\pm$ 0.002 & 0.583 $\pm$ 0.004 & \textbf{0.745} $\pm$ 0.005 & 0.676 $\pm$ 0.004 & \textbf{0.896} $\pm$ 0.021 & 0.772 $\pm$ 0.017 \\ 

\noalign{\vskip 0.25ex}\cdashline{2-11}\noalign{\vskip 0.75ex}

& Augmentation & \textbf{0.487} $\pm$ 0.007 & \textbf{0.469} $\pm$ 0.003 & \textbf{0.437} $\pm$ 0.003 & \textbf{0.595} $\pm$ 0.005 & 0.748 $\pm$ 0.007 & 0.694 $\pm$ 0.005 & \textbf{0.927} $\pm$ 0.019 & 0.784 $\pm$ 0.018 \\ 

\midrule
\midrule

& & \multicolumn{2}{c}{Reasoning} & \multicolumn{5}{c}{Classification} & All \\

\cmidrule(l{2pt}r{2pt}){3-4} \cmidrule(l{2pt}r{2pt}){5-9} \cmidrule(l{2pt}r{2pt}){10-10} 

\textbf{LCLMs} & \textbf{Methods} & Tracking7 & Web & Banking77 & DialogRE & Discovery & FewNERD & GoEmotion & Average \\

\midrule
\midrule

\multirow{3}{*}{\textbf{Gemini Pro}} 

& Random & 0.294 $\pm$ 0.029 & 0.675 $\pm$ 0.021 & 0.878 $\pm$ 0.002 & 0.661 $\pm$ 0.009 & 0.195 $\pm$ 0.007 & 0.568 $\pm$ 0.012 & 0.393 $\pm$ 0.007 & 0.574 $\pm$ 0.010 \\ 

& Best Selection & 0.311 $\pm$ 0.031 & \textbf{0.700} $\pm$ 0.028 & \textbf{0.886} $\pm$ 0.004 & \textbf{0.709} $\pm$ 0.014 & 0.204 $\pm$ 0.011 & 0.569 $\pm$ 0.006 & \textbf{0.413} $\pm$ 0.006 & 0.586 $\pm$ 0.011 \\ 

\noalign{\vskip 0.25ex}\cdashline{2-11}\noalign{\vskip 0.75ex}

& Augmentation & 0.307 $\pm$ 0.031 & \textbf{0.768} $\pm$ 0.040 & \textbf{0.889} $\pm$ 0.004 & \textbf{0.698} $\pm$ 0.010 & \textbf{0.209} $\pm$ 0.009 & 0.574 $\pm$ 0.008 & \textbf{0.428} $\pm$ 0.006 & \textbf{0.601} $\pm$ 0.012 \\ 

\midrule
\midrule

& & \multicolumn{6}{c}{Translation} & \multicolumn{2}{c}{Reasoning} \\

\cmidrule(l{2pt}r{2pt}){3-8} \cmidrule(l{2pt}r{2pt}){9-10} 

\textbf{LCLMs} & \textbf{Methods} & ENG to BEM & ENG to KMR & ENG to EWE & ENG to SPA & ENG to FRA & ENG to DEU & Date & Salient \\

\midrule
\midrule

\multirow{3}{*}{\textbf{Gemini Flash}} 

& Random & 0.419 $\pm$ 0.006 & 0.427 $\pm$ 0.004 & 0.363 $\pm$ 0.002 & 0.573 $\pm$ 0.004 & 0.726 $\pm$ 0.004 & 0.666 $\pm$ 0.005 & 0.754 $\pm$ 0.022 & 0.682 $\pm$ 0.019 \\ 

& Best Selection & 0.421 $\pm$ 0.002 & 0.434 $\pm$ 0.002 & 0.360 $\pm$ 0.003 & 0.575 $\pm$ 0.002 & 0.732 $\pm$ 0.003 & 0.673 $\pm$ 0.001 & 0.777 $\pm$ 0.030 & 0.687 $\pm$ 0.015 \\ 

\noalign{\vskip 0.25ex}\cdashline{2-11}\noalign{\vskip 0.75ex}

& Augmentation & \textbf{0.436} $\pm$ 0.006 & \textbf{0.460} $\pm$ 0.002 & \textbf{0.378} $\pm$ 0.004 & \textbf{0.594} $\pm$ 0.007 & 0.737 $\pm$ 0.010 & 0.676 $\pm$ 0.012 & \textbf{0.804} $\pm$ 0.037 & \textbf{0.714} $\pm$ 0.013 \\ 

\midrule
\midrule

& & \multicolumn{2}{c}{Reasoning} & \multicolumn{5}{c}{Classification} & All \\

\cmidrule(l{2pt}r{2pt}){3-4} \cmidrule(l{2pt}r{2pt}){5-9} \cmidrule(l{2pt}r{2pt}){10-10} 

\textbf{LCLMs} & \textbf{Methods} & Tracking7 & Web & Banking77 & DialogRE & Discovery & FewNERD & GoEmotion & Average \\

\midrule
\midrule

\multirow{3}{*}{\textbf{Gemini Flash}} 

& Random & 0.256 $\pm$ 0.030 & 0.582 $\pm$ 0.033 & 0.868 $\pm$ 0.004 & 0.541 $\pm$ 0.008 & 0.065 $\pm$ 0.007 & 0.521 $\pm$ 0.006 & 0.362 $\pm$ 0.016 & 0.520 $\pm$ 0.011 \\ 

& Best Selection & 0.270 $\pm$ 0.031 & 0.566 $\pm$ 0.031 & 0.872 $\pm$ 0.006 & 0.547 $\pm$ 0.012 & \textbf{0.083} $\pm$ 0.007 & \textbf{0.532} $\pm$ 0.002 & \textbf{0.385} $\pm$ 0.006 & 0.528 $\pm$ 0.010 \\ 

\noalign{\vskip 0.25ex}\cdashline{2-11}\noalign{\vskip 0.75ex}

& Augmentation & 0.281 $\pm$ 0.035 & 0.609 $\pm$ 0.040 & \textbf{0.880 }$\pm$ 0.006 & \textbf{0.578} $\pm$ 0.025 & \textbf{0.090} $\pm$ 0.005 & \textbf{0.537} $\pm$ 0.009 & \textbf{0.392} $\pm$ 0.015 & \textbf{0.544} $\pm$ 0.015 \\ 

\bottomrule

\end{tabular}
}
\vspace{-0.1in}
\end{table*}

\section{Augmenting ICL Demonstrations to Increase Context Capacity of LCLMs}

\subsection{ICL Example Augmentation Approach}
Recall that recent advances in LCLMs offer unprecedented context capacity, potentially amplifying ICL performance by including more examples. However, the available examples sometimes fall short of filling this expanded capacity, and this under-utilization of the context may result in suboptimal performance. To address this, we introduce a simple yet effective ICL sample augmentation approach designed to increase the context capacity of LCLMs, while being scalable for many-shot scenarios. This method consists of synthetic example generation and low-quality example filtering.

\paragraph{Generation of Synthetic Examples}
Formally, let $\mathcal{D} = \{(\vx_i, \vy_i)\}_{i=1}^k$ be a set of available ICL examples for a target task. The objective is to generate a set of synthetic examples $\mathcal{D}' = \{(\vx'_j, \vy'_j)\}_{j=1}^m$ (to supplement the original dataset $\mathcal{D}$), such that the augmented set of examples $\mathcal{D}_{\texttt{AUG}} = \mathcal{D} \cup \mathcal{D}'$ can increase the utilization of the available context capacity of LCLMs. To operationalize this, we generate each synthetic example ($\vx'_j, \vy'_j$) by prompting an LM with randomly selected real examples from $\mathcal{D}$ as context, to ensure the generated data retains meaningful characteristics relevant to the task.

\paragraph{Filtering Out Low-Quality Examples}
Once the synthetic examples are generated, we filter out low-quality instances that may introduce noise or irrelevant information. To do this, we design a function $f$ that assigns a quality score to each synthetic example $(\vx'_j, \vy'_j)$ based on its contextual relevance and alignment with real examples as well as overall quality. Specifically, each synthetic example is rated on a 5-point Likert scale by prompting the LM 30 times with the synthetic and 30 real examples. We then compute an aggregate score using a weighted average of scores with their corresponding probabilities from the LM. Only the synthetic examples that exceed the quality threshold, $\tau$, are retained in the augmented example set, as follows:
\begin{equation}
    \mathcal{D}_{\texttt{AUG}} = \mathcal{D} \cup \{(\vx'_j, \vy'_j) \; | \; f(\vx'_j, \vy'_j, \mathcal{D}) \geq \tau \}_{j=1}^m. \nonumber
\end{equation}
Notably, our data augmentation process is efficient, as it is performed offline and does not contribute to inference-time overhead. Also, it takes under 10 seconds per example, which can be done in parallel.

\subsection{Experimental Setup}
For synthetic data generation and filtering, we use Gemini Pro, one of the state-of-the-art LMs. We focus on tasks that underutilize the context capacity of LCLMs even when all available samples are provided, such as translation, reasoning, and classification. For each task, we generate 3,000 examples and retain only those with a quality score above the median among the generated samples. As a result, we use the original examples and 1,500 synthetic examples. The prompts used to elicit data generation and filtering are provided in Appendix~\ref{appendix:prompt}.

\subsection{Experimental Results}

\paragraph{Main Results}
As shown in Table~\ref{tab:augmentation}, which compares the example augmentation approach (with random selection) to other sample selection strategies, the augmentation approach demonstrates substantial performance gains across various datasets, which can be attributed to the greater diversity and volume of ICL examples achieved through synthetic data generation, leading to the effective utilization of the context capacity of LCLMs. Also, like the random selection approach, our augmentation method allows the reuse of the same examples across all queries. Thus, due to key-value caching, the augmentation approach is as efficient as random selection while achieving superior performance.

\begin{figure*}[t!]
    \centering
    \begin{minipage}{0.635\textwidth}
        \centering
        \begin{minipage}{\textwidth}
            \centering
            \includegraphics[width=0.32\textwidth]{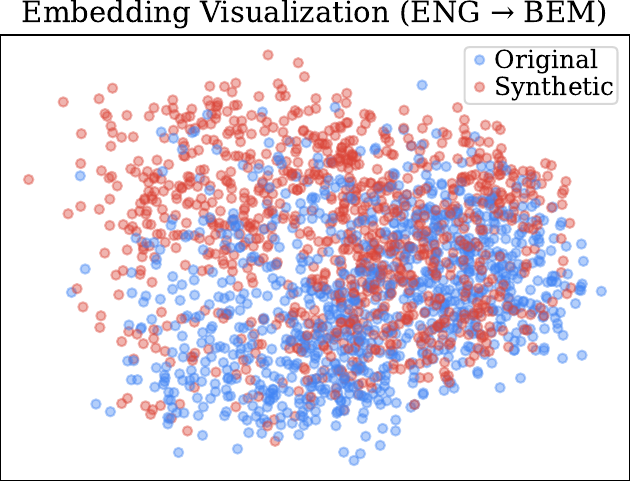}
            \includegraphics[width=0.32\textwidth]{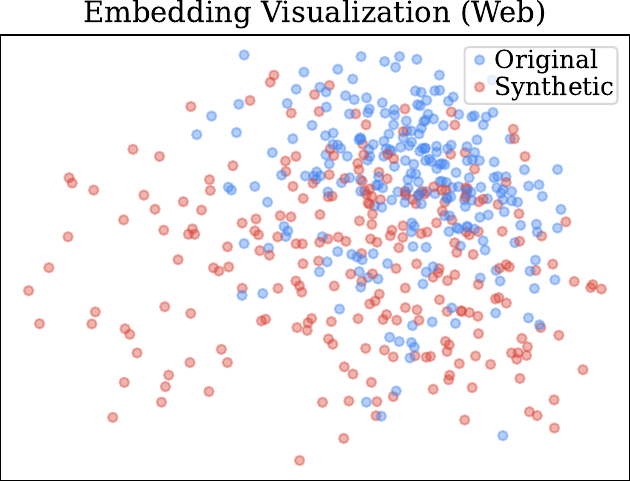}
            \includegraphics[width=0.32\textwidth]{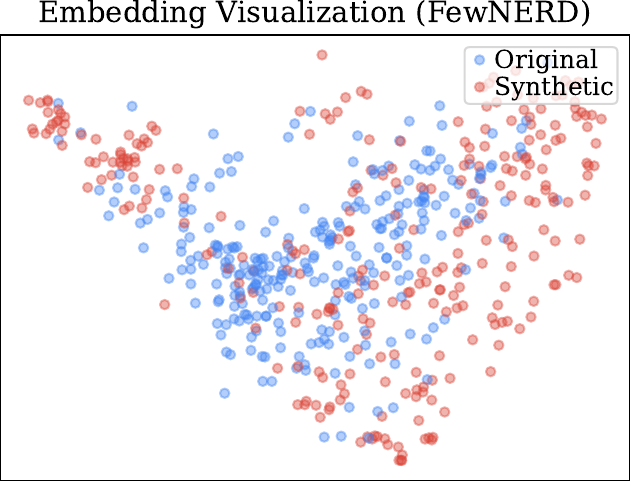}
            \vspace{-0.1in}
            \caption{Visualization of embedding-space with original and synthetic examples.}
            \label{fig:embedding-space}
        \end{minipage}
    \end{minipage}
    \hfill
    \begin{minipage}{0.355\linewidth}
        \centering
        \vspace{-0.025in}
        \captionof{table}{Results on Similarity (embedding-level similarity between original and synthetic examples) and Volume (relative expansion of the convex hull with augmented examples).}
        \vspace{-0.1in}
        \resizebox{0.975\textwidth}{!}{
        \renewcommand{\arraystretch}{0.6}
        \renewcommand{\tabcolsep}{4.0mm}
        \begin{tabular}{lcc}
        \toprule
        Tasks & Similarity & Volume \\
        \midrule
        Translation & 0.5715 & 1.6563 \\
        Reasoning & 0.8099 & 3.2328 \\
        Classification & 0.6252 & 2.7931 \\
        \bottomrule
        \end{tabular}
        }
        \label{tab:augmentation-similarity}
    \end{minipage}
\end{figure*}

\begin{figure*}[t!]
    \centering
    \begin{minipage}{0.54\textwidth}
        \centering
        \begin{minipage}{0.48\textwidth}
            \centering
            \includegraphics[width=\textwidth]{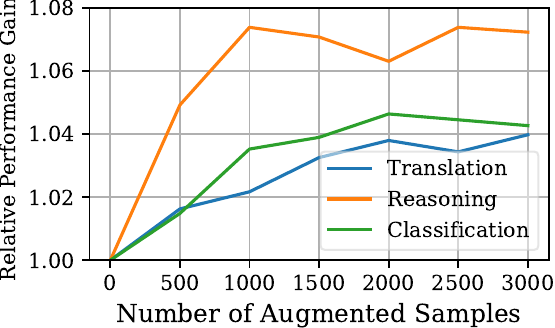}
        \end{minipage}
        \begin{minipage}{0.48\textwidth}
            \centering
            \includegraphics[width=\textwidth]{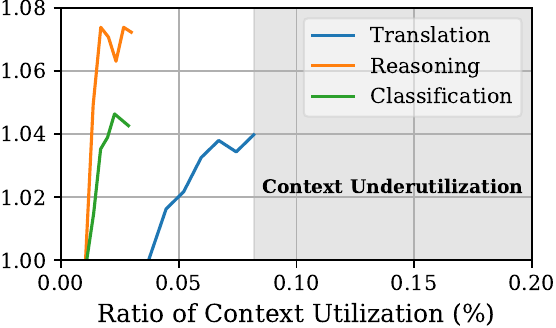}
        \end{minipage}
        \vspace{-0.1in}
        \caption{Results with augmented examples according to the size of synthetic samples (Left) and context utilization of Gemini Pro (Right).}
        \label{fig:augmentation_varying}
    \end{minipage}
    \hfill
    \begin{minipage}{0.45\linewidth}
        \centering
        \captionof{table}{Results on ablation study, where w/o Filtering and w/o Original denote results based on augmented samples without filtering and without original samples, respectively. Only Original shows results without augmentation.}
        \vspace{-0.1in}
        \label{tab:ablation}
        \small
        \centering
        \resizebox{0.975\textwidth}{!}{
        \renewcommand{\arraystretch}{0.68}
        \begin{tabular}{lccc}
        \toprule
        
        Methods & Translation & Reasoning & Classification \\
        
        \midrule
        \midrule
        
        Augmentation & \textbf{0.571} $\pm$ 0.005 & \textbf{0.696} $\pm$ 0.027 & \textbf{0.560} $\pm$ 0.008 \\
        
        \noalign{\vskip 0.25ex}\cdashline{1-4}\noalign{\vskip 0.75ex}
        
        w/o Filtering & 0.552 $\pm$ 0.005 & 0.666 $\pm$ 0.031 & 0.548 $\pm$ 0.009 \\
        
        w/o Original & 0.544 $\pm$ 0.002 & 0.611 $\pm$ 0.025 & 0.531 $\pm$ 0.007 \\
        
        \noalign{\vskip 0.25ex}\cdashline{1-4}\noalign{\vskip 0.75ex}
        
        Only Original & 0.553 $\pm$ 0.004 & 0.650 $\pm$ 0.023 & 0.539 $\pm$ 0.007 \\
        
        \bottomrule
        
        \end{tabular}
        }
    \end{minipage}
    \vspace{-0.075in}
\end{figure*}

\paragraph{Analysis on Augmented Data}
Beyond performance improvements, we analyze the characteristics of the augmented data to better understand its impact on ICL. First, as visualized in Figure~\ref{fig:embedding-space}, the embedding-space distribution of augmented examples closely follows that of real examples while expanding the overall data coverage, which suggests that the synthetic examples effectively capture task-relevant features without deviating substantially from the original data distribution. In addition, we further quantify this expansion through two metrics: the similarity between original and synthetic examples, and the relative expansion of the convex hull with augmented examples compared to that formed by original examples, and report results in Table~\ref{tab:augmentation-similarity}. From this, we observe that while synthetic examples maintain a high degree of similarity to real examples (ensuring alignment with the task), they also significantly increase the volume of the data distribution. This balance between relevance and diversity highlights why our augmentation approach effectively enhances ICL performance. 

\begin{figure*}[t]
    \centering
    \includegraphics[width=0.99\linewidth]{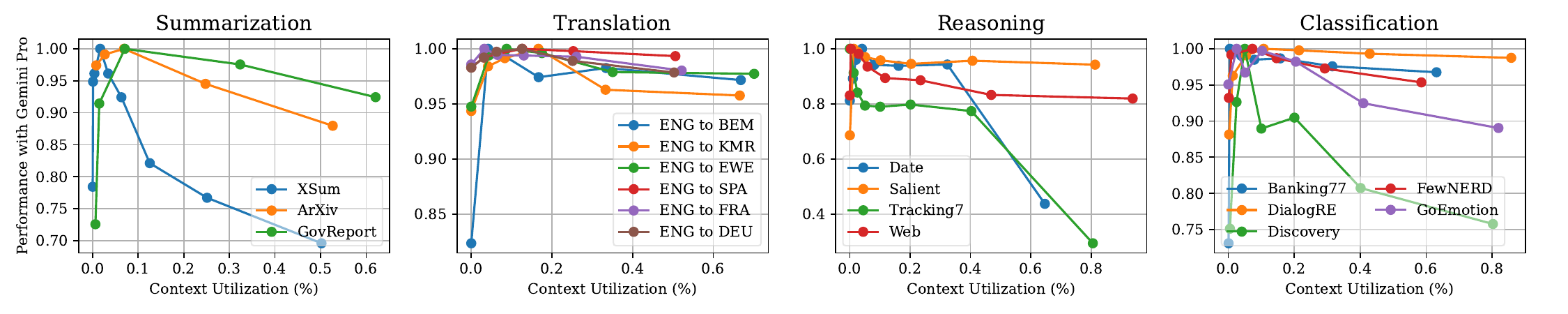}
    \vspace{-0.18in}
    \caption{Results across different percentages of context size utilized in LCLMs, where the x-axis represents the percentage of the full LCLM context used (according to the number of tokens over the full token length), and the y-axis shows the relative performance compared to the highest performance achieved for each dataset. Results are averaged over multiple runs.}
    \label{fig:long}
    \vspace{-0.1in}
\end{figure*}

Finally, we analyze the impact of the number of augmented examples on performance and their corresponding context utilization in LCLMs. As shown in Figure~\ref{fig:augmentation_varying}, while increasing the number of synthetic examples initially improves performance, it eventually plateaus, indicating diminishing returns. Also, despite augmentation improving context utilization, we find that even at peak performance, the augmented data occupies less than 3\% of the full context capacity of LCLMs, which is significantly below the scale that LCLMs can handle (Figure~\ref{fig:long}). These suggest an interesting future work to develop more advanced augmentation strategies to increase the context utilization of LCLMs.

\paragraph{Ablation Study on Augmentation}
To see how each component in the augmentation approach contributes to performance gains, we conduct an ablation study. As shown in Table~\ref{tab:ablation}, we observe that the full augmentation method (called Augmentation), which uses both original and filtered synthetic examples, achieves the best performance. In contrast, when the filtering step is omitted, performance decreases, indicating that filtering contributes positively by removing lower-quality examples. Also, a large performance drop occurs when original samples are excluded from the augmented set. This suggests that although filtering helps maintain quality, the synthetic samples generated still do not match the quality of the original examples. Thus, while our augmentation approach is effective, further research could improve data generation techniques to improve the quality of the synthetic examples.

\begin{figure*}[th!]
    \centering
    \includegraphics[width=0.99\linewidth]{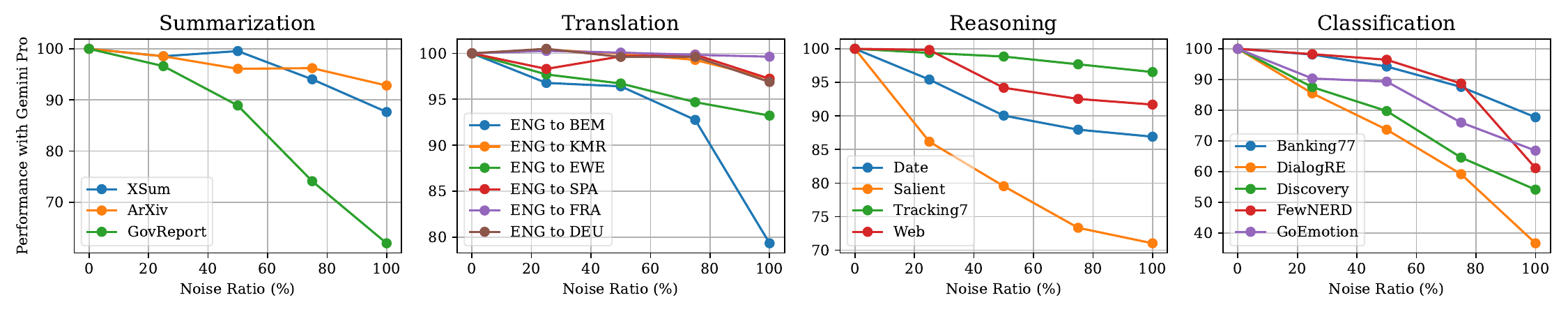}
    \vspace{-0.18in}
    \caption{Results with varying the ratio of noisy examples within the context of LCLMs, where we report the relative performance over the ICL without noisy examples (i.e., the noise ratio of 0) and the results are averaged over multiple runs.}
    \label{fig:noise}
    \vspace{-0.05in}
\end{figure*}

\section{Behaviors of LCLM-Enabled ICL}

\subsection{LCLM-Based ICL with Noisy Examples}
LCLMs can accommodate a large number of diverse ICL examples, which raises the question of the impact and risk of including noisy examples in the context. We investigate how the performance of LCLM-enabled ICL is impacted when some or all of the ICL examples are noisy. To simulate noisy examples, we modify the outputs of a subset of in-context demonstrations by replacing their outputs with outputs from other randomly selected demonstrations. As shown in Figure~\ref{fig:noise}, LCLM-enabled ICL is largely robust to noise when the proportion of noisy examples is relatively low (i.e., below 25\%). This observation highlights why augmented examples, even if slightly lower quality, can still enhance performance as it increases the utilization of the context window. In contrast, when the amount of noise exceeds this threshold, LCLMs become vulnerable to the negative effects of noise and the performance notably declines. This adverse effect is more pronounced for challenging tasks, such as low-resource translation (e.g., English to Bemba or Ewe). This is likely because LCLMs are less familiar with those tasks, and therefore rely more on learning from in-context examples.

\subsection{LCLM-Based ICL with Long Context}
\label{sec:long_context}
As the context length capacity of LCLMs continues to grow, it becomes increasingly important to assess whether LCLMs can reliably utilize a large number of ICL examples. To investigate this, we conduct an experiment analyzing the performance as a function of the context utilization. Specifically, we gradually increase the number of examples by powers of two, and if the entire set of examples within the dataset is used, we further extend the context utilization by repeating these examples. The hypothesis being tested is that if LCLMs can effectively understand and utilize extremely long context, performance should remain consistent even with repeated examples, as the presence of duplicates should not impact contextual understanding. However, as shown in Figure~\ref{fig:long}, a substantial performance decline occurs when LCLMs are pushed to use extremely large contexts. Specifically, this decline generally begins when more than 25\% of the available context capacity is utilized. Also, the performance drop is pronounced in tasks such as xsum, which requires generating abstractive summaries (unlike other summarization datasets like arXiv or GovReport) and in tasks demanding complex reasoning such as date understanding (Date) and object tracking (Tracking7). These findings suggest that while LCLMs can handle moderately long contexts, they encounter limitations with exceedingly large contexts, particularly in tasks requiring fine-grained reasoning or abstractive generation. This may be due to challenges in distinguishing and integrating relevant information across numerous examples, especially when tasks require high levels of nuanced abstraction and precise reasoning.

\section{Related Work}

\paragraph{LCLMs}
The field of language modeling has witnessed remarkable advancements with Language Models (LMs)~\cite{GPT3, GPT-4, gemini, llama3}. However, earlier LMs were constrained by relatively short context windows, typically handling only a few thousand tokens at a time, which limits their applicability in advanced tasks requiring broader context comprehension, such as document-level summarization or complex reasoning~\cite{longsumm, BBH}. To address this, recent efforts have led to the development of LCLMs, designed to process much larger contexts, sometimes accommodating over a million tokens within a single prompt~\cite{gemini}. To mention a few, models like Longformer and BigBird~\cite{Longformer, BigBird} incorporate sparse attention mechanisms to efficiently handle extended contexts without compromising on computational feasibility. Also, LongRoPE extends the context window of LMs to 2M tokens by interpolating their specific positional embeddings~\cite{LongRoPE}.

\paragraph{In-Context Learning}
In-Context Learning (ICL) is a recent paradigm that enables language models to learn from examples provided within their input context and then perform given tasks~\cite{GPT3, ICL/1, ICL/2}. Since its introduction, previous studies have concentrated on developing the strategies to optimize the quality and arrangement of in-context examples to maximize performance, especially given the limitations of early LMs on context length. For example, these approaches include selecting examples that maximize relevance to the target query~\cite{ICL/KNN/1, ICL/KNN/2}, ensuring diversity among examples to cover a range of possible cases~\cite{ICL/diversity/1, ICL/diversity/2}, strategically ordering examples to improve model adaptation~\cite{order/1, order/2}, and prioritizing examples by their ease of learning based on their difficultly~\cite{ICL/easy/1, ICL/easy/2}. Yet, as the context capacity expands with LCLMs, these conventional selection strategies warrant re-evaluation, particularly in many-shot settings; thus, we focus on revisiting them.

\paragraph{Many-Shot ICL}
Early approaches in many-shot ICL have primarily focused on the paradigm shift brought by the ability to incorporate a larger number of examples in-context~\cite{ManyShotICL/1, ManyShotICL/2}, without giving much consideration to example selection strategies. Such many-shot ICL methods have demonstrated performance comparable to fine-tuning. Also, there is a very recent work that explores retrieval strategies in many-shot ICL~\cite{ManyShotICL/2}; however, they use models with relatively limited context capacities (e.g., under 100k tokens with Llama 2), resulting in restrictions on the number of examples included and, consequently, making retrieval-based methods appear more advantageous. However, contrary to this finding, we uncover that this advantage diminishes as the context capacity increases, allowing random sampling to perform on par with more sophisticated selection methods when a large number of examples is used. Lastly, other recent efforts include establishing benchmarks for long-context ICL~\cite{loft, LongICLBench}. Unlike prior studies, our work offers a novel perspective by systematically re-evaluating traditional selection strategies in the expanded context regime and highlighting the shift from selection optimization to effectively leveraging the extensive context space in many-shot ICL, with data augmentation.

\section{Conclusion}
\vspace{-0.03in}
We explored ICL in the context of LCLMs, investigating whether traditional sample selection strategies remain effective in many-shot scenarios and observing that they offer minimal to zero performance gains over simple random selection. We also highlighted the emerging challenge of underutilized context in low-resource tasks due to limited example availability, and proposed a data augmentation strategy, which substantially boosts performance by increasing context utilization of LCLMs. Lastly, we analyzed the behavior of LCLM-enabled ICL when operating with extremely long context and in the presence of noisy examples, and found that while performance improves with added examples, it plateaus and even declines when the context becomes too long, with increased vulnerability to noise in complex tasks. This suggests promising future directions in making LCLMs more robust to lengthy context and noise examples alongside the direction of extending their context length.

\vspace{-0.03in}
\section*{Limitations}
\vspace{-0.03in}
While this work explores the new opportunity of ICL with LCLMs, a couple of limitations can be considered. First, the computational cost associated with LCLMs remains a significant challenge, particularly for researchers and practitioners in resource-constrained settings. Second, while the proposed data augmentation method enhances context utilization of LCLMs and improves ICL performance, the quality of synthetic examples often falls short of the quality of original data. Addressing them through cost-efficient strategies for leveraging LCLMs and developing improved data augmentation techniques would be an exciting area for future work. Lastly, a theoretical understanding of why LCLMs exhibit insensitivity to example selection in many-shot settings remains an open research question.

\vspace{-0.03in}
\section*{Ethics Statement}
\vspace{-0.03in}
We believe this work does not raise any direct ethical concerns, as it primarily focuses on advancing the understanding of ICL with LCLMs. However, as with any other application of LCLM-based ICL, careful consideration must be given to the quality of the examples used in the context. Specifically, the inclusion of biased, harmful, or otherwise problematic examples in the input context can propagate or amplify these issues in the model's outputs, and we advise practitioners to carefully evaluate and select ICL examples to avoid potential issues.

\bibliography{custom}

\clearpage
\appendix

\section{Prompts}
\label{appendix:prompt}

We provide the prompts used for many-shot ICL on translation, summarization, and reasoning tasks in Table~\ref{tab:prompt:ICL} and on classification tasks in Table~\ref{tab:prompt:classification}. Also, we provide the prompts used for synthetic data augmentation and filtering in Table~\ref{tab:prompt:augmentation}.

\section{Detailed Experimental Setup}
\label{appendix:setup}

\paragraph{Configuration}
For all experiments, we use the default hyperparameters for Gemini and Llama. 

\paragraph{Ratio of Augmented Data}
We use original examples alongside 1,500 synthetic samples (filtered from an initial set of 3,000 examples according to their quality scores); therefore, the percentage of augmented samples varies depending on the size of the original examples in each dataset. Specifically, for the translation task where there are around 1,000 original examples, synthetic samples comprise around 60\% of the total examples. For reasoning tasks (having around 100 to 150 examples), synthetic samples constitute 90-94\% of the total examples. For the classification task (e.g., Banking77 dataset), with 385 original examples, synthetic samples account for around 80\% of the total examples.

\section{Detailed Experimental Results}

\paragraph{Results with CoT}
It is worth noting that while developing the approach to better utilize many examples within the expanded context windows of LCLMs with advanced prompting techniques, such as Chain-of-Thought (CoT)~\cite{CoT}, represents an orthogonal but promising future research direction, as an initial foray into this area, we perform experiments with CoT on the reasoning task (as it may benefit from explicit step-by-step thinking procedures) and report results in Table~\ref{tab:cot}. From this, we then observe that the CoT prompting strategy improves the performance on most datasets (except for Date whose performance is already high without CoT), demonstrating that there may be a potential to enhance the performance of LCLM-enabled many-shot ICL via advanced prompting.

\begin{figure*}[t]
    \centering
    \vspace{-0.1in}
    \includegraphics[width=0.99\linewidth]{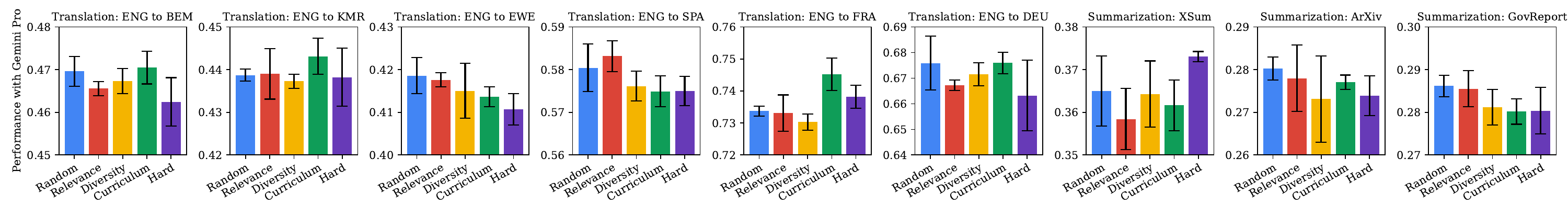}

    \vspace{-0.1in}
    
    \includegraphics[width=0.99\linewidth]{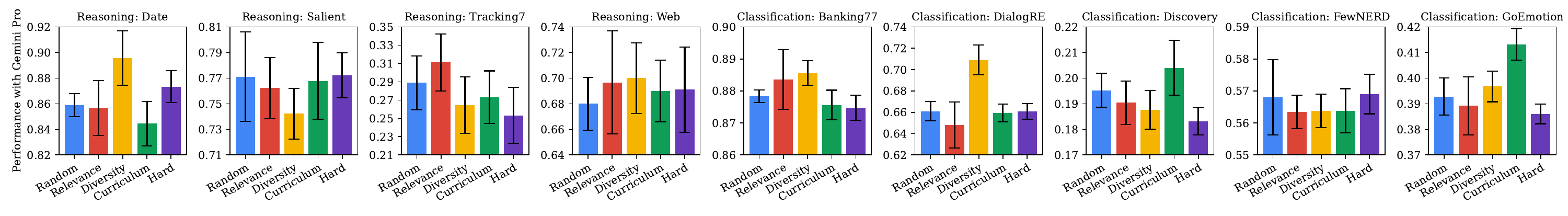}

    \vspace{-0.2in} % Adjust spacing if needed
    \rule{\linewidth}{0.5pt} % Adds a horizontal line

    \includegraphics[width=0.99\linewidth]{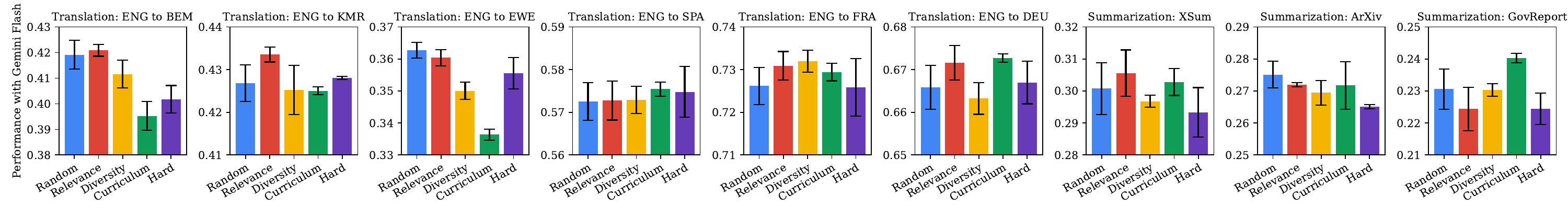}

    \vspace{-0.1in}

    \includegraphics[width=0.99\linewidth]{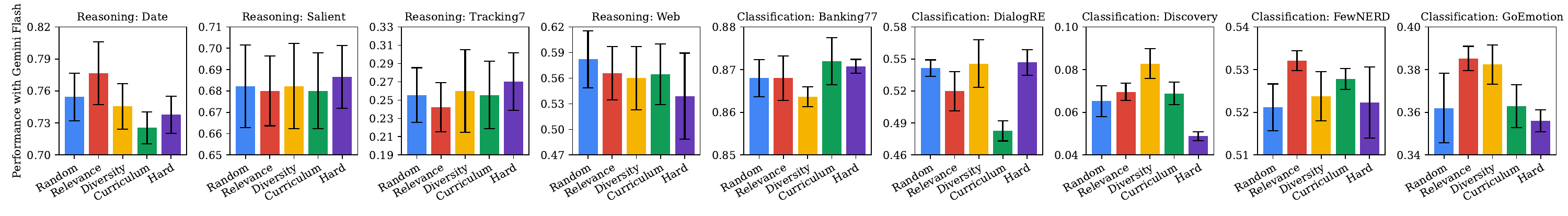}

    \vspace{-0.2in} % Adjust spacing if needed
    \rule{\linewidth}{0.5pt} % Adds a horizontal line

    \includegraphics[width=0.99\linewidth]{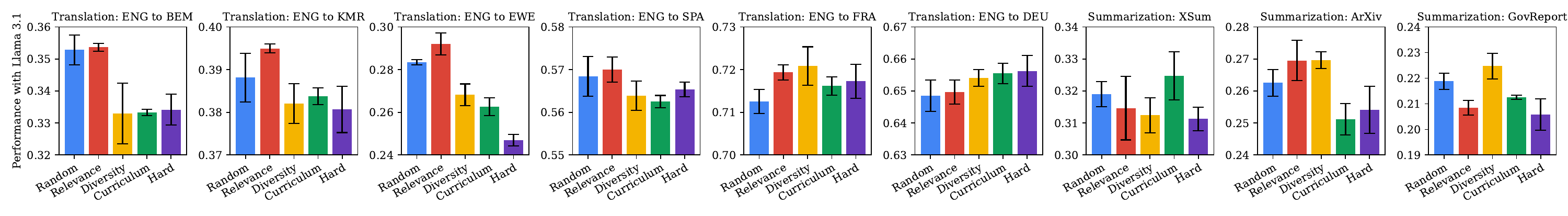}

    \vspace{-0.1in}

    \includegraphics[width=0.99\linewidth]{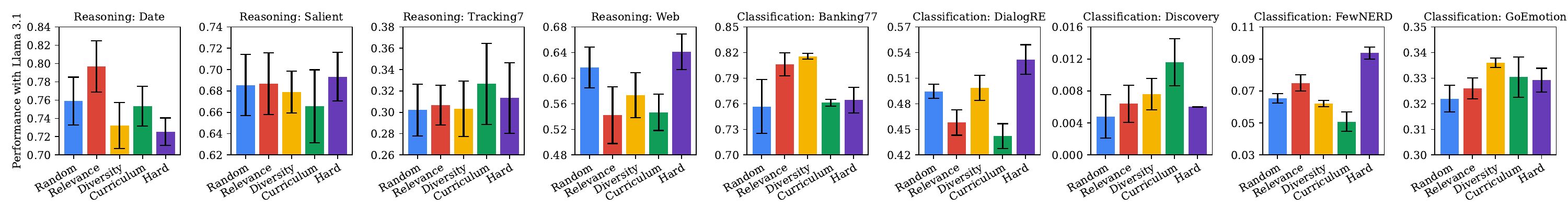}
    
    \vspace{-0.17in}
    \caption{Detailed results of various sample selection approaches on ICL with LCLMs, such as Gemini Pro (Top), Gemini Flash (Middle), and Llama 3.1 (Bottom), across four different tasks (translation, summarization, reasoning, and extreme classification) with 18 datasets. Each bar represents the averaged performance, with the upper and lower limits indicating standard deviation.}
    \label{fig:selection}
    \vspace{-0.18in}
\end{figure*}
\begin{table}[t]
\caption{Results of LCLM-enabled ICL on reasoning datasets with and without Chain-of-Thought (CoT)~\cite{CoT}.}
\vspace{-0.125in}
\label{tab:cot}
\small
\centering
\resizebox{0.475\textwidth}{!}{
\renewcommand{\arraystretch}{1.125}
\begin{tabular}{lcccc}
\toprule

\textbf{Methods} & \textbf{Date} & \textbf{Salient} & \textbf{Tracking7} & \textbf{Web} \\

\midrule
\midrule

Many-Shot ICL & 0.927 & 0.784 & 0.307 & 0.768 \\

Many-Shot ICL with CoT & 0.918 & 0.810 & 0.334 & 0.771 \\

\bottomrule

\end{tabular}
}
\vspace{-0.in}
\end{table}

\begin{table*}[t]
    \small
    \centering
    \caption{A list of prompts that we use for many-shot ICL on translation, summarization, and reasoning tasks.}
    \label{tab:prompt:ICL}
    \vspace{-0.1in}
    \resizebox{1\textwidth}{!}{
    \renewcommand{\arraystretch}{1.5}
        \begin{tabular}{ll}
        \toprule
        \multicolumn{1}{p{.14\textwidth}}{\textbf{Types}} & \makecell{\multicolumn{1}{p{.86\textwidth}}{\textbf{Prompts}}} \\
        \midrule
        \midrule
        \multicolumn{1}{p{.14\textwidth}}{Translation} & 
        \makecell{
            \multicolumn{1}{p{.86\textwidth}}{You are an expert translator. I am going to give you one or more example pairs of text snippets where the first is in \{SOURCE\_LANGUAGE\} and the second is a translation of the first snippet into \{TARGET\_LANGUAGE\}.}\\
            \multicolumn{1}{p{.86\textwidth}}{}\\
            \multicolumn{1}{p{.86\textwidth}}{The sentences will be written as the following format: }\\
            \multicolumn{1}{p{.86\textwidth}}{\{SOURCE\_LANGUAGE\}: <first sentence>}\\
            \multicolumn{1}{p{.86\textwidth}}{\{TARGET\_LANGUAGE\}: <translated first sentence>}\\
            \multicolumn{1}{p{.86\textwidth}}{}\\
            \multicolumn{1}{p{.86\textwidth}}{After the example pairs, I am going to provide another sentence in \{SOURCE\_LANGUAGE\} and I want you to translate it into \{TARGET\_LANGUAGE\}. Give only the translation, and no extra commentary, formatting, or chattiness. Translate the text from \{SOURCE\_LANGUAGE\} to \{TARGET\_LANGUAGE\}.}\\
            \multicolumn{1}{p{.86\textwidth}}{}\\
            \multicolumn{1}{p{.86\textwidth}}{\{EXAMPLES\}}\\
            \multicolumn{1}{p{.86\textwidth}}{}\\
            \multicolumn{1}{p{.86\textwidth}}{\{TARGET\_QUERY\}}\\
        }\\
        \noalign{\vskip 0.25ex}\cdashline{1-2}\noalign{\vskip 0.75ex}
        \multicolumn{1}{p{.14\textwidth}}{Summarization} & 
        \makecell{
            \multicolumn{1}{p{.86\textwidth}}{You are an expert in article summarization. I am going to give you one or more example pairs of article and its summary in fluent English.}\\
            \multicolumn{1}{p{.86\textwidth}}{}\\
            \multicolumn{1}{p{.86\textwidth}}{The pairs will be written as the following format: }\\
            \multicolumn{1}{p{.86\textwidth}}{Article: <article>}\\
            \multicolumn{1}{p{.86\textwidth}}{Summary: <summary>}\\
            \multicolumn{1}{p{.86\textwidth}}{}\\
            \multicolumn{1}{p{.86\textwidth}}{After the example pairs, I am going to provide another article and I want you to summarize it. Give only the summary, and no extra commentary, formatting, or chattiness.}\\
            \multicolumn{1}{p{.86\textwidth}}{}\\
            \multicolumn{1}{p{.86\textwidth}}{\{EXAMPLES\}}\\
            \multicolumn{1}{p{.86\textwidth}}{}\\
            \multicolumn{1}{p{.86\textwidth}}{\{TARGET\_QUERY\}}\\
        } \\
        \noalign{\vskip 0.25ex}\cdashline{1-2}\noalign{\vskip 0.75ex}
        \multicolumn{1}{p{.14\textwidth}}{Reasoning} & 
        \makecell{
            \multicolumn{1}{p{.86\textwidth}}{You are an expert in multiple-choice question answering tasks. I am going to give you one or more example pairs of question and its answer in a multiple-choice question answering format.}\\
            \multicolumn{1}{p{.86\textwidth}}{}\\
            \multicolumn{1}{p{.86\textwidth}}{The pairs will be written as the following format: }\\
            \multicolumn{1}{p{.86\textwidth}}{Question: <question>}\\
            \multicolumn{1}{p{.86\textwidth}}{Answer: <answer>}\\
            \multicolumn{1}{p{.86\textwidth}}{}\\
            \multicolumn{1}{p{.86\textwidth}}{After the example pairs, I am going to provide another question and I want you to predict its answer. Give only the answer that follows a consistent format as in the provided examples, and no extra commentary, formatting, or chattiness.}\\
            \multicolumn{1}{p{.86\textwidth}}{}\\
            \multicolumn{1}{p{.86\textwidth}}{\{EXAMPLES\}}\\
            \multicolumn{1}{p{.86\textwidth}}{}\\
            \multicolumn{1}{p{.86\textwidth}}{\{TARGET\_QUERY\}}\\
        } \\
        \bottomrule
        \end{tabular}
    }
\end{table*}

\begin{table*}[t]
    \small
    \centering
    \caption{A list of prompts that we use for many-shot ICL on five different extreme classification tasks.}
    \label{tab:prompt:classification}
    \vspace{-0.1in}
    \resizebox{1\textwidth}{!}{
    \renewcommand{\arraystretch}{1.5}
        \begin{tabular}{ll}
        \toprule
        \multicolumn{1}{p{.14\textwidth}}{\textbf{Types}} & \makecell{\multicolumn{1}{p{.86\textwidth}}{\textbf{Prompts}}} \\
        \midrule
        \midrule
        \multicolumn{1}{p{.14\textwidth}}{BANKING77} & 
        \makecell{
            \multicolumn{1}{p{.86\textwidth}}{I am going to give you one or more example pairs of customer service query and its intent. }\\
            \multicolumn{1}{p{.86\textwidth}}{}\\
            \multicolumn{1}{p{.86\textwidth}}{The pairs will be written as the following format: }\\
            \multicolumn{1}{p{.86\textwidth}}{service query: <query>}\\
            \multicolumn{1}{p{.86\textwidth}}{intent category: <category>}\\
            \multicolumn{1}{p{.86\textwidth}}{}\\
            \multicolumn{1}{p{.86\textwidth}}{After the example pairs, I am going to provide another customer service query and I want you to classify the label of it that must be one among the intent categories provided in the examples. Give only the category, and no extra commentary, formatting, or chattiness.}\\
            \multicolumn{1}{p{.86\textwidth}}{}\\
            \multicolumn{1}{p{.86\textwidth}}{\{EXAMPLES\}}\\
            \multicolumn{1}{p{.86\textwidth}}{}\\
            \multicolumn{1}{p{.86\textwidth}}{\{TARGET\_QUERY\}}\\
        }\\
        \noalign{\vskip 0.25ex}\cdashline{1-2}\noalign{\vskip 0.75ex}
        \multicolumn{1}{p{.14\textwidth}}{DialogRE} & 
        \makecell{
            \multicolumn{1}{p{.86\textwidth}}{I am going to give you one or more examples of the dialogue, the list of entity pairs within it, and their corresponding relation types.}\\
            \multicolumn{1}{p{.86\textwidth}}{}\\
            \multicolumn{1}{p{.86\textwidth}}{The examples will be written as the following format: }\\
            \multicolumn{1}{p{.86\textwidth}}{Dialogue: <dialogue>}\\
            \multicolumn{1}{p{.86\textwidth}}{The list of k entity pairs are (<entity 1>, <entity 2>), ...}\\
            \multicolumn{1}{p{.86\textwidth}}{The k respective relations between each entity pair are: <relation>, ...}\\
            \multicolumn{1}{p{.86\textwidth}}{}\\
            \multicolumn{1}{p{.86\textwidth}}{After the examples, I am going to provide another dialogue along with its associated entity pairs, and I want you to classify their corresponding relation types that must be one among the relation types provided in the examples. Give only the relations, and no extra commentary, formatting, or chattiness.}\\
            \multicolumn{1}{p{.86\textwidth}}{}\\
            \multicolumn{1}{p{.86\textwidth}}{\{EXAMPLES\}}\\
            \multicolumn{1}{p{.86\textwidth}}{}\\
            \multicolumn{1}{p{.86\textwidth}}{\{TARGET\_QUERY\}}\\
        } \\
        \noalign{\vskip 0.25ex}\cdashline{1-2}\noalign{\vskip 0.75ex}
        \multicolumn{1}{p{.14\textwidth}}{Discovery} & 
        \makecell{
            \multicolumn{1}{p{.86\textwidth}}{I am going to give you one or more example pairs of two sentences and the conjunction word between them.}\\
            \multicolumn{1}{p{.86\textwidth}}{}\\
            \multicolumn{1}{p{.86\textwidth}}{The pairs will be written as the following format: }\\
            \multicolumn{1}{p{.86\textwidth}}{<sentence 1> ( ) <sentence 2> }\\
            \multicolumn{1}{p{.86\textwidth}}{the most suitable conjunction word in the previous ( ) is  <conjunction word>}\\
            \multicolumn{1}{p{.86\textwidth}}{}\\
            \multicolumn{1}{p{.86\textwidth}}{After the example pairs, I am going to provide another two sentences and I want you to classify the conjunction word between them that must be one among the conjunction words provided in the examples. Give only the conjunction word, and no extra commentary, formatting, or chattiness.}\\
            \multicolumn{1}{p{.86\textwidth}}{}\\
            \multicolumn{1}{p{.86\textwidth}}{\{EXAMPLES\}}\\
            \multicolumn{1}{p{.86\textwidth}}{}\\
            \multicolumn{1}{p{.86\textwidth}}{\{TARGET\_QUERY\}}\\
        } \\
        \noalign{\vskip 0.25ex}\cdashline{1-2}\noalign{\vskip 0.75ex}
        \multicolumn{1}{p{.14\textwidth}}{FewNERD} & 
        \makecell{
            \multicolumn{1}{p{.86\textwidth}}{I am going to give you one or more examples of the sentence, the named entities within it, and their corresponding entity types.}\\
            \multicolumn{1}{p{.86\textwidth}}{}\\
            \multicolumn{1}{p{.86\textwidth}}{The examples will be written as the following format: }\\
            \multicolumn{1}{p{.86\textwidth}}{Sentence: <sentence> }\\
            \multicolumn{1}{p{.86\textwidth}}{<named entity>: <entity type> }\\
            \multicolumn{1}{p{.86\textwidth}}{}\\
            \multicolumn{1}{p{.86\textwidth}}{After the example pairs, I am going to provide another comment and I want you to classify the label of it that must be one among the emotion categories provided in the examples. Give only the category, and no extra commentary, formatting, or chattiness.}\\
            \multicolumn{1}{p{.86\textwidth}}{}\\
            \multicolumn{1}{p{.86\textwidth}}{\{EXAMPLES\}}\\
            \multicolumn{1}{p{.86\textwidth}}{}\\
            \multicolumn{1}{p{.86\textwidth}}{\{TARGET\_QUERY\}}\\
        } \\
        \noalign{\vskip 0.25ex}\cdashline{1-2}\noalign{\vskip 0.75ex}
        \multicolumn{1}{p{.14\textwidth}}{GoEmotion} & 
        \makecell{
            \multicolumn{1}{p{.86\textwidth}}{I am going to give you one or more example pairs of comment and its emotion category.}\\
            \multicolumn{1}{p{.86\textwidth}}{}\\
            \multicolumn{1}{p{.86\textwidth}}{The pairs will be written as the following format: }\\
            \multicolumn{1}{p{.86\textwidth}}{comment: <comment> }\\
            \multicolumn{1}{p{.86\textwidth}}{emotion category: <category> }\\
            \multicolumn{1}{p{.86\textwidth}}{}\\
            \multicolumn{1}{p{.86\textwidth}}{After the example pairs, I am going to provide another sentence, and I want you to classify the named entities within it and their corresponding entity types that must be one among the entity types provided in the examples. Give only the named entities and their corresponding entity types, and no extra commentary, formatting, or chattiness.}\\
            \multicolumn{1}{p{.86\textwidth}}{}\\
            \multicolumn{1}{p{.86\textwidth}}{\{EXAMPLES\}}\\
            \multicolumn{1}{p{.86\textwidth}}{}\\
            \multicolumn{1}{p{.86\textwidth}}{\{TARGET\_QUERY\}}\\
        } \\
        \bottomrule
        \end{tabular}
    }
\end{table*}

\begin{table*}[t]
    \small
    \centering
    \caption{A list of prompts that we use for generating synthetic demonstrations and filtering them of low-quality.}
    \label{tab:prompt:augmentation}
    \vspace{-0.1in}
    \resizebox{1\textwidth}{!}{
    \renewcommand{\arraystretch}{1.5}
        \begin{tabular}{ll}
        \toprule
        \multicolumn{1}{p{.14\textwidth}}{\textbf{Types}} & \makecell{\multicolumn{1}{p{.86\textwidth}}{\textbf{Prompts}}} \\
        \midrule
        \midrule
        \multicolumn{1}{p{.14\textwidth}}{Generation} & 
        \makecell{
            \multicolumn{1}{p{.86\textwidth}}{You are an expert in data augmentation. You will be provided with a series of demonstrations that show how a task is performed. Your objective is to generate a new example that closely follows the pattern, structure, and style of the demonstrations. Carefully analyze the key steps, transitions, and output style in the provided demonstrations. Then, create a new sample that maintains consistency in format and correctness while introducing variety in content.}\\
            \multicolumn{1}{p{.86\textwidth}}{}\\
            \multicolumn{1}{p{.86\textwidth}}{Here are the demonstrations: }\\
            \multicolumn{1}{p{.86\textwidth}}{}\\
            \multicolumn{1}{p{.86\textwidth}}{\{EXAMPLES\}}\\
            \multicolumn{1}{p{.86\textwidth}}{}\\
            \multicolumn{1}{p{.86\textwidth}}{Now, as an expert, generate a new sample that aligns with the original demonstrations:}\\
        } \\
        \noalign{\vskip 0.25ex}\cdashline{1-2}\noalign{\vskip 0.75ex}
        \multicolumn{1}{p{.14\textwidth}}{Filtering} & 
        \makecell{
            \multicolumn{1}{p{.86\textwidth}}{You are an expert in assessing data quality. Given the original set of samples, your task is to carefully evaluate the provided sample in comparison to the original samples. Based on your expertise, determine whether the provided sample is of high quality, meeting or exceeding the standards set by the original set. }\\
            \multicolumn{1}{p{.86\textwidth}}{}\\
            \multicolumn{1}{p{.86\textwidth}}{Here are the original samples: }\\
            \multicolumn{1}{p{.86\textwidth}}{\{EXAMPLES\}}\\
            \multicolumn{1}{p{.86\textwidth}}{}\\
            \multicolumn{1}{p{.86\textwidth}}{Now, as an expert, evaluate the provided sample:}\\
            \multicolumn{1}{p{.86\textwidth}}{\{GENERATED\_SAMPLE\}}\\
            \multicolumn{1}{p{.86\textwidth}}{}\\
            \multicolumn{1}{p{.86\textwidth}}{Please provide only a single numerical rating (1, 2, 3, 4, or 5) based on the quality of the sample, without any additional commentary, formatting, or chattiness.}\\
        } \\
        \bottomrule
        \end{tabular}
    }
\end{table*}

\end{document}